\documentclass[11pt,a4paper]{article}
\usepackage[hyperref]{acl2021}
\usepackage{times}
\usepackage{latexsym}

\usepackage{balance}

\usepackage{microtype}

\usepackage{amssymb}
\usepackage{pifont}

\usepackage{pbox}
\usepackage{hyperref}
\usepackage{url}
\usepackage{comment}
\usepackage[flushleft]{threeparttable}
\usepackage{graphicx}
\usepackage{amsmath}
\usepackage{tabularx}
\usepackage{multirow, makecell}
\usepackage{algorithm}
\usepackage{xcolor}
\usepackage{booktabs}
\usepackage{paralist}
\usepackage{subfig}
\usepackage{caption}

\usepackage{amsthm}
\theoremstyle{definition}
\newtheorem{definition}{Task}

\DeclareMathOperator*{\softmax}{softmax}

\usepackage[noabbrev]{cleveref}
\crefname{section}{§}{§§}
\Crefname{section}{§}{§§}

\aclfinalcopy 
\def\aclpaperid{3155} 

\definecolor{MyColor}{RGB}{50, 100, 250}
\definecolor{Orange}{RGB}{244, 101, 66}
\definecolor{Red}{RGB}{255, 0, 0}
\definecolor{Green}{RGB}{0, 255, 0}
\definecolor{Blue}{RGB}{0, 0, 255}

\newcommand{\mytilde}{\raise.17ex\hbox{$\scriptstyle\mathtt{\sim}$}}

\title{Select, Extract and Generate: Neural Keyphrase Generation \\ with Layer-wise Coverage Attention}

\author{
Wasi Uddin Ahmad$^\dagger$\thanks{~~Work done during internship at Yahoo Research.}, Xiao Bai$^\ddagger$, Soomin Lee$^\ddagger$, Kai-Wei Chang$^\dagger$ \\
$^\dagger$University of California, Los Angeles, $^\ddagger$Yahoo Research \\
$^\dagger$\texttt{\{wasiahmad,kwchang\}@cs.ucla.edu} \\ $^\ddagger$\texttt{\{xbai,soominl\}@verizonmedia.com}
}

\date{}

\begin{document}
\maketitle

\begin{abstract}
Natural language processing techniques have demonstrated promising results in keyphrase generation. However, one of the major challenges in \emph{neural} keyphrase generation is processing long documents using deep neural networks. Generally, documents are truncated before given as inputs to neural networks. Consequently, the models may miss essential points conveyed in the target document. To overcome this limitation, we propose \emph{SEG-Net}, a neural keyphrase generation model that is composed of two major components, (1) a selector that selects the salient sentences in a document and (2) an extractor-generator that jointly extracts and generates keyphrases from the selected sentences. SEG-Net uses Transformer, a self-attentive architecture, as the basic building block with a novel \emph{layer-wise} coverage attention to summarize most of the points discussed in the document. The experimental results on seven keyphrase generation benchmarks from scientific and web documents demonstrate that SEG-Net outperforms the state-of-the-art neural generative methods by a large margin.
\end{abstract}

\section{Introduction}
\label{sec:intro}

Keyphrases are short pieces of text that summarize the key points discussed in a document.
They are useful for many natural language processing and information retrieval tasks \cite{wilson-etal-2005-recognizing, berend-2011-opinion, tang2017qalink, subramanian-etal-2018-neural, zhang2017mike, wan2008single, jones1999phrasier, kim-etal-2013-applying, hulth-megyesi-2006-study, hammouda2005corephrase, wu2008keyword, dave2010pattern}.
In the automatic keyphrase generation task, the input is a document, and the output is a set of keyphrases that can be categorized as \emph{present} or \emph{absent} keyphrases.
Present keyphrases appear exactly in the target document, while absent keyphrases are only semantically related and have partial or no overlap to the target document.
We provide an example of a target document and its keyphrases in Figure \ref{table:example}.

\begin{figure}
\centering
\vspace{3mm}
\resizebox{\linewidth}{!}{%
\begin{tabular}{ p{1.0\linewidth}}
\hline
\textbf{Title:} 
[1] {\color{blue}natural language processing} {\color{red}technologies} for developing a language learning environment . \\
\hline
\textbf{Abstract:}
[1] so far , {\color{blue}computer assisted language learning} ( call ) comes in many different flavors . [1] our research work focuses on developing an {\color{blue} integrated e learning} environment that allows improving language skills in specific contexts . [1] {\color{blue} integrated e learning} environment means that it is a {\color{red}web} based solution \ldots , for instance , {\color{red}web} browsers or email clients . [0] it should be accessible \ldots [1] {\color{blue} natural language processing} ( nlp ) forms the technological basis for developing such a {\color{red}learning} framework . [0] the paper gives an overview \ldots [0] therefore , on the one hand , it explains creation \ldots [0] on the other hand , it describes existing nlp standards . [0] based on our requirements , the paper gives \ldots [1] \ldots necessary developments in e {\color{red}learning} to keep in mind . \\
\hline
\textbf{Present:} 
{\color{blue}natural language processing}; {\color{blue}computer assisted language learning}; {\color{blue}integrated e learning} \\\hline
\textbf{Absent:} 
{\color{red}semantic web technologies}; {\color{red}learning of foreign languages}   \\
\hline 
\end{tabular}
}
\vspace{-2mm}
\caption{
Example of a document with present and absent keyphrases. The value (0/1) in brackets ([]) represent sentence \emph{salience} label.
}
\label{table:example}
\vspace{-2mm}
\end{figure}

In recent years, the neural sequence-to-sequence (Seq2Seq) framework \cite{sutskever2014sequence} has become the fundamental building block in keyphrase generation models.
Most of the existing approaches \cite{meng-etal-2017-deep,chen-etal-2018-keyphrase,yuan-etal-2020-one,chen2019guided} adopt the Seq2Seq framework with attention \cite{luong-etal-2015-effective,bahdanau2014neural} and copy mechanism \cite{see-etal-2017-get,gu-etal-2016-incorporating}.
However, present phrases indicate the indispensable segments of a target document. 
Emphasizing on those segments improves document understanding that can lead a model to coherent absent phrase generation. This motivates to jointly model keyphrase extraction and generation \cite{chen-etal-2019-integrated}.

To generate a comprehensive set of keyphrases, reading the complete target document is necessary.
However, to the best of our knowledge, none of the previous neural methods read the full content of a document as it can be thousands of words long.
Existing models truncate the target document; take the first few hundred words as input and ignore the rest of the document that may contain salient information.
On the contrary, a significant fraction of a long document may not associate with the keyphrases. 
Presumably, selecting the salient segments from the target document and then predicting the keyphrases from them would be effective.

To address the aforementioned challenges, in this paper, we propose SEG-Net (stands for \textbf{S}elect, \textbf{E}xtract, and \textbf{G}enerate) that has two major components, (1) a \emph{sentence-selector} that selects the salient sentences in a document, and (2) an \emph{extractor-generator} that predicts the present keyphrases and generates the absent keyphrases jointly.
The motivation to design the sentence-selector is to decompose a long target document into a list of sentences, and identify the salient ones for keyphrase generation.
We consider a sentence as salient if it contains present keyphrases or overlaps with absent keyphrases. 
As shown in Figure \ref{table:example}, we split the document into a list of sentences and classify them with salient and non-salient labels.
A similar notion is adopted in prior works on text summarization \cite{chen-bansal-2018-fast,lebanoff-etal-2019-scoring} and question answering \cite{min-etal-2018-efficient}.
We employ \emph{Transformer} \cite{vaswani2017attention} as the backbone of the extractor-generator in SEG-Net.

We equip the extractor-generator with a novel \emph{layer-wise} coverage attention such that the generated keyphrases summarize the entire target document.
The layer-wise coverage attention keeps track of the target document segments that are covered by previously generated phrases to guide the self-attention mechanism in Transformer while attending the encoded target document in future generation steps.
We evaluate SEG-Net on five benchmarks from scientific articles and two benchmarks from web documents to demonstrate its effectiveness over the state-of-the-art neural generative methods.
We perform ablation and analysis to show that selecting salient sentences improve present keyphrase extraction and the layer-wise coverage attention and facilitates absent keyphrase generation.
Our novel contributions are as follows.
\begin{compactenum}
    \item SEG-Net that identifies the salient sentences in the target document first and then use them to generate a set of keyphrases.
    \item A layer-wise coverage attention.
\end{compactenum}

\section{Problem Definition}
\label{sec:prob_definition}

Keyphrase generation task is defined as given a text document $x$, generate a set of keyphrases $\mathcal{K} = \{k^1, k^2, \ldots, k^{|\mathcal{K}|}\}$ where the document $x = [x_1, \ldots, x_{|x|}]$ and each keyphrase $k^i = [k^i_1, \ldots, k^i_{|k^i|}]$ is a sequence of words.
A text document can be split into a list of sentences, $\mathcal{S}_x = [s^1_x, s^2_x, \ldots, s^{|{S}|}_x]$ where each sentence $s^i_x = [x_j, \ldots, x_{j+|s^i|-1}]$ is a consecutive subsequence of the document $x$ with begin index $j \leq |x|$ and end index $(j+|s^i|) < |x|$.
In literature, keyphrases are categorized into two types, \emph{present} and \emph{absent}.
A present keyphrase is a consecutive subsequence of the document, while an absent keyphrase is not.
However, an absent keyphrase may have a partial overlapping with the document's word sequence.
We denote the sets of present and absent keyphrases as $\mathcal{K}_p = \{k^{1}_p, k^{2}_p, \ldots, k_p^{|\mathcal{K}^p|}\}$ and $\mathcal{K}_a = \{k_a^{1}, k_a^{2}, \ldots, k_a^{|\mathcal{K}^a|}\}$, respectively.
Hence, we can express a set of keyphrases as $\mathcal{K} = \mathcal{K}_p \cup \mathcal{K}_a$.

SEG-Net decomposes the keyphrase generation task into three sub-tasks. 
We define them below. 

\theoremstyle{definition}

\vspace{1mm}
\begin{definition}[\textbf{Salient Sentence Selection}]
\label{def:def1}
Given a list of sentences $\mathcal{S}_x$, predict a binary label ($0/1$) for each sentence $s^i_x$. The label $1$ indicates that the sentence contains a present keyphrase or overlaps with an absent keyphrase.
The output of the selector is a list of salient sentences $\mathcal{S}^{sal}_x$.
\end{definition}

\vspace{1mm}
\begin{definition}[\textbf{Present Keyphrase Extraction} ]
\label{def:def2}
Given $\mathcal{S}^{sal}_x$ as a concatenated sequence of words, predict a label (B/I/O) for each word that indicates if it is a constituent of a present keyphrase. 
\end{definition}

\vspace{1mm}
\begin{definition}[\textbf{Absent Keyphrase Generation}]
\label{def:def3}
Given $\mathcal{S}^{sal}_x$ as a concatenated sequence of words, generate a concatenated sequence of keyphrases in a sequence-to-sequence fashion.
\end{definition}

\section{SEG-Net for Keyphrase Generation}
\label{sec:model}
Our proposed model, SEG-Net jointly learns to extract and generate present and absent keyphrases from the salient sentences in a target document.
The key advantage of SEG-Net is the maximal utilization of the information from the input text in order to generate a set of keyphrases that summarize all the key points in the target document.
SEG-Net consists of a \emph{sentence-selector} and an \emph{extractor-generator}.
The sentence-selector identifies the salient sentences from the target document (Task \ref{def:def1}) that are fed to the extractor-generator to predict both the present and absent keyphrases (Task \ref{def:def2}, \ref{def:def3}).
We detail them in this section.

\subsection{Embedding Layer} 
\label{sec:emb_layer}
The embedding layer maps each word in an input sequence to a low-dimensional vector space.
We train three embedding matrices, $W_e, W_{pos},$ and $W_{seg}$ that convert a word, its absolute position, and segment index into vector representations of size $d_{model}$.
The segment index of a word indicates the index of the sentence that it belongs to.
In addition, we obtain a character-level embedding for each word
using Convolutional Neural Networks (CNN) \cite{kim-2014-convolutional}.
To learn a fixed-length vector representation of a word, we add the four embedding vectors element-wise.
To form the vector representations of the keyphrase tokens, we only use their word and character-level embeddings.

\subsection{Sentence-Selector} 
The objective of the sentence-selector is to predict the salient sentences in a document, as described in Task \ref{def:def1}.
Given a sentence, $s^i_x = [x_j, \ldots, x_{j+|s^i|-1}]$ from a document $x$, the selector predicts the salience probability of that input sentence.
First, the embedding layer maps each word in the sentence into a $d_{model}$ dimensional vector.
The sequence of word vectors are fed to a stack of Transformer encoder layers that produce a sequence of output representations $[o_j, \ldots, o_{j+|s^i|-1}]$ where $o_t \in R^{d_{model}}$.
Then we apply max and mean pooling on the output representations to form $s_{max}, s_{mean} \in R^{d_{model}}$ that are concatenated $s_{pool} = s_{max} \oplus s_{mean}$ to form the sentence embedding vector.
We feed the vector $s_{pool}$ through a three-layer, batch-normalized \cite{ioffe2015batch} maxout network \cite{goodfellow2013maxout} to predict the salience probability.

\begin{figure}[!ht]
\centering
\includegraphics[width=1.0\linewidth]{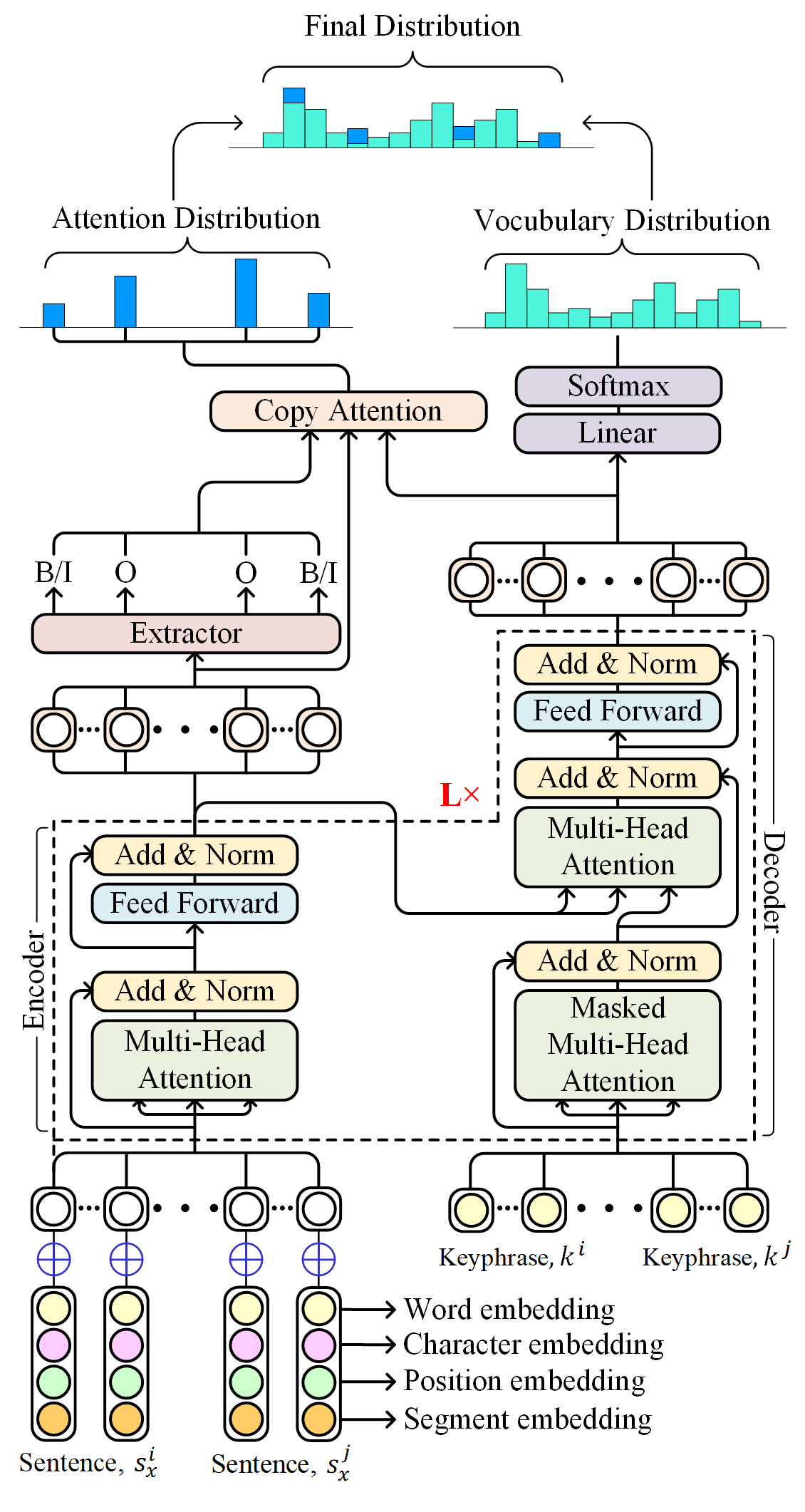}
\caption{
Overview of the Extractor-Generator module of SEG-Net. The major components are encoder, extractor, and decoder. The encoder encodes the salient sentences of the input document. The extractor predicts the present keyphrase's constituent words while the decoder generates the absent keyphrases word by word.
}
\label{fig:reader}
\end{figure}

\subsection{Extractor-Generator}
The extractor-generator module in SEG-Net takes a list of salient sentences from a document as an input that are concatenated to form a sequence of words and predicts the present and absent keyphrases.
We illustrate the extractor-generator module in Figure \ref{fig:reader} and describe its major components as follows.

\paragraph{Encoder}
The encoder consists of an embedding layer followed by an $L$-layer  Transformer encoder.
Each word in the input sequence $[x_1, \ldots, x_n]$ is first mapped to an embedding vector.
Then the sequence of word embeddings is fed to the Transformer encoder that produces contextualized word representations $[o_1^l, \ldots, o_n^l]$ where $l=1, \ldots, L$ using the multi-head self-attention mechanism.


\paragraph{Extractor}
In a nutshell, the extractor acts as a 3-way classifier that predicts a tag for each word in the BIO format.
The extractor takes $[o_1^L, \ldots, o_n^L]$ as input and predicts the probability of each word being a constituent of a present keyphrase.
\begin{equation*}
    p_j = \softmax \big(W_{r_2}(\tanh(W_{r_1} o_j^L + b_{r_1})) + b_{r_2} \big),
\end{equation*}
where $W_{r_1}, W_{r_2}, b_{r_1}, b_{r_2}$ are trainable parameters.

\paragraph{Decoder}
The decoder generates the absent keyphrases as a concatenated sequence of words $[y_1^\ast, \ldots, y_m^\ast]$ where $m$ is the sum of the length of the phrases.
The decoder predicts the absent phrases word by word given previously predicted words in a greedy fashion.
The decoder employs an embedding layer, $L$-layers of Transformer decoder followed by a softmax layer.
The embedding layer converts the words into vector representations that are fed to the Transformer decoder.
We use relative positional encoding \cite{shaw-etal-2018-self} to inject order information of the keyphrase terms.
The output of the last ($L$-th) decoder layer $h_1^{L}, \ldots, h_m^{L}$ is passed through a softmax layer to predict a probability distribution over the vocabulary $V$.
\begin{equation}
\label{eq:softmax}
    p(y_{t}^\ast|y_{1:t-1}^\ast, x) = \softmax(W_{v} h^{L}_{t} + b_{v}),
\end{equation}
where $W_{v} \in R^{|V| \times d_{model}}$ and $b_{word} \in R^{|V|}$.


\paragraph{Coverage Attention}
The coverage attention \cite{tu-etal-2016-modeling,yuan-etal-2020-one,chen-etal-2018-keyphrase} keeps track of the parts in the document that has been covered by previously generated phrases and encourages future generation steps to summarize the other segments of the target document.
The underlying idea is to decay the attention weights of the previously attended input tokens while decoder attends the encoded input tokens at time step, $t$.
To equip the multi-layer structure of the Transformer with a \emph{layer-wise} coverage attention, we adopt the layer-wise encoder-decoder attention technique \cite{he2018layer}.
We compute the attention weights, $\alpha_{ti} = \frac{e'_{ti}}{\sum^{n}_{k=1} e'_{tk}}$ 
in encoder-decoder attention at each layer where $e'_{ti}$ is as follows.
\begin{gather}
\label{eq:mulhead}
e'_{ti} = 
\begin{cases}
    \exp(e_{ti}) & \text{if } t = 1 \\
    \frac{\exp(e_{ti})}{\sum_{k=1}^{t-1} \exp(e_{ki})} & \text{otherwise},
\end{cases}
\end{gather}
where $e_{ti}$ is the scaled-dot product between the target token $y_t$ and the input token $x_i$.


\paragraph{Copy Attention}
Absent keyphrases have partial or no overlapping with the target document.
With the copy mechanism, we want the decoder to learn to copy phrase terms that overlap with the target document.
Hence, we adopt the copying mechanism and use an additional attention layer to learn the copy distribution on top of the decoder stack.

Formally, we take the output from the last layer of the encoder $[o_1^L, \ldots, o^L_n]$ and compute the attention score of the decoder output $h_t^L$ at time step $t$ as: $att(o^L_i, h^L_t) = o^L_i W_{att} h^L_t$.
Then we compute the context vector, $c_t^L$ at time step t:
\begin{equation*}
    a_{ti}^L = \frac{att(o_i^L, h_t^L)}{\sum_{k=1}^n \exp(att(o_k^L, h_t^L))} ;\ c_t^L = \sum_{i=1}^n a_{ti}^L o_i^L.
\end{equation*}
The copy mechanism uses the attention weights $a_{ti}^L$ as the probability distribution $P(y_{t}^\ast = x_i | u_t = 1) = a_{ti}^L$ to copy the input tokens $x_i$.
We compute the probability of using the copy mechanism at the decoding step t as $p(u_t = 1) = \sigma (W_u[h_t^L || c_t^L] + b_u)$, where $||$ denotes the vector concatenation operator.
Then we obtain the final probability distribution for the output token $y_t^\ast$ as: $P(y_{t}^\ast) = P(u_t = 0) P(y_{t}^\ast|u_t = 0) + P(u_t = 1) P(y_{t}^\ast|u_t = 1)$
where $P(y_{t}^\ast|u_t = 0)$ is defined in Eq. \eqref{eq:softmax}. 
All probabilities are conditioned on $y_{1:t-1}^\ast, x$, but we omit them to keep the notations simple.


\subsection{Learning Objectives}
\label{sec:learning}
We individually train the sentence-selector and the extractor-generator in SEG-Net.

\paragraph{Sentence-Selector}
For each sentence in a document $x$, the selector predicts the salience label. 
We choose the sentences containing present keyphrases or overlap with absent keyphrases as the gold salient sentences and use the weighted cross-entropy loss for selector training.
\begin{equation}
\label{eq:cross_entropy}
    \mathcal{L}_s = -\frac{1}{|x|} \sum_{j=1}^{|x|} \omega \vartheta_j^{\ast}\log\vartheta_j + (1 - \vartheta_j^{\ast})\log(1 - \vartheta_j),
\end{equation}
where $\vartheta_j^{\ast} \in \{0, 1\}$ is the ground-truth label for the $j$-th sentence and $\omega$ is a hyper-parameter to balance the importance of salient and non-salient sentences.

\paragraph{Extractor-Generator}
The extractor-generator takes a list of salient sentences as a concatenated sequence of words.
For each word of the input sequence, the extractor predicts whether the word appears in a contiguous subsequence that matches a present keyphrase.
The extractor treats the task as a binary classification task and we compute the extraction loss $\mathcal{L}_e$ as in Eq. \eqref{eq:cross_entropy}.

\begin{table*}[t]
\centering
\resizebox{\linewidth}{!}{%
\begin{tabular}{l|c|c|c|c|c|c|c|c}
\hline
Dataset & \# Example  & \begin{tabular}{@{}c@{}}Max / Avg. \\ Source Len. \end{tabular} & \begin{tabular}{@{}c@{}}Max / Avg. \\ \# Sentence \end{tabular} & \% Sent$^{\star}$ & \begin{tabular}{@{}c@{}}Max / Avg. \\ Kp Len. \end{tabular} & \begin{tabular}{@{}c@{}} Avg. \\ \# Kp \end{tabular} & \% PKp & \% AKp \\
\hline
KP20k & 20,000 & 1,438 / 179.8 & 108 / 7.8 & 29.2 & 23 / 2.04 & 5.28 & 62.9 & 37.1 \\
Inspec & 500 & 386 / 128.7 & 23 / 5.5 & 16.5 & 10 / 2.48 & 9.83 & 73.6 & 26.4 \\
Krapivin & 400 & 554 / 182.6 & 28 / 8.2 & 28.3 & 6 / 2.21 & 5.84 & 55.7 & 44.3 \\
Nus & 211 & 973 / 219.1 & 42 / 11.8 & 32.6 & 70 / 2.22 & 11.65 & 54.4 & 45.6 \\
SemEval & 100 & 473 / 234.8 & 22 / 11.9 & 27.0 & 11 / 2.38 & 14.66 & 42.6 & 57.4 \\
KPTimes & 20,000 & 7,569 / 777.9 & 631 / 28.9 & 35.4 & 18 / 1.84 & 5.27 & 58.8 & 41.2 \\
In-house & 26,000 & 9,745 / 969.1 & 538 / 35.6 & 44.0 & 16 / 2.69 & 4.08 & 37.5 & 62.5 \\
\hline
\end{tabular}
}
\caption{Summary of the test portion of the keyphrase benchmarks used in experiments. Sent$^{\star}$ represents the percentage of non-salient sentences in the input text.
\% PKp and \% AKp indicate the percentage of present and absent keyphrases, respectively. 
}
\label{table:statitics}
\end{table*}

The decoder in extractor-generator generates the list of absent keyphrases in a sequence-to-sequence fashion.
We compute the negative log-likelihood $\mathcal{L}_{g}$ of the ground-truth keyphrases.
\begin{equation}
\mathcal{L}_{g} = -\sum_{t=1}^{n} \log\ p(y_t^{\ast} | y_{1}^{\ast}, \ldots, y_{t-1}^{\ast}, x),
\end{equation}
where $n$ is sum of the length of all absent phrases.
The overall loss to train the extractor-generator is computed as a weighted average of the extraction and generation loss, $\mathcal{L}_{eg} = \beta\mathcal{L}_{e} + (1 - \beta)\mathcal{L}_{g}$.

\section{Experiment Setup}

\subsection{Datasets and Preprocessing}

We conduct experiments on five scientific benchmarks from the computer science domain: KP20k \cite{meng-etal-2017-deep}, Inspec \cite{hulth-2003-improved}, Krapivin \cite{krapivin2009large}, NUS \cite{nguyen2007keyphrase}, and SemEval \cite{kim-etal-2010-semeval}.
Each example from these datasets consists of the title, abstract, and a list of keyphrases.
Following previous works \cite{meng-etal-2017-deep,chan-etal-2019-neural,chen2019guided,chen-etal-2019-integrated,yuan-etal-2020-one}, we use the training set of the largest dataset, KP20k, to train and employ the testing datasets from all the benchmarks to evaluate the baselines and our models.
KP20k dataset consists of 530,000 and 20,000 articles for training and validation, respectively.
We remove all the articles from the training portion of KP20k that overlaps with its validation set, or in any of the five testing sets. 
After filtering, the KP20k dataset contains 509,818 training examples that we use to train all the baselines and our models.

We perform experiments on two web-domain datasets that consist of news articles and general web documents.
The first dataset is KPTimes \cite{gallina-etal-2019-kptimes} that provides news text paired with editor-curated keyphrases.
The second dataset is an \emph{in-house} dataset generated from the click logs of a large-scale commercial web search engine. 
Specifically, we randomly sampled web documents that were clicked at least once during the month of February in 2019. 
For each sampled web document, we collected 20 queries that led to the highest number of clicks on it. 
This design choice is motivated by the observation that queries frequently leading to clicks on a web document usually summarize the main concepts in the document. 
We further filter out the less relevant queries by ranking them based on the number of clicks. 
The relevance score for each query is assigned by an in-house query-document relevance model.
We also remove duplicate queries by comparing their bag-of-words representation.\footnote{We perform stemming before computing the bag-of-words representations.} 
The dataset consists of 206,000, 24,000, and 26,000 unique web documents for training, validation, and evaluation, respectively. 

Statistics of the test portion of the experiment datasets are provided in Table \ref{table:statitics} in Appendix.
Following \citet{meng-etal-2017-deep}, we apply lowercasing, tokenization and replacing digits with $\langle digit \rangle$ symbol to preprocess all the datasets.
We use spaCy \cite{spacy} for tokenization and collecting the sentence boundaries.

\begin{table*}[ht]
\centering
\resizebox{\linewidth}{!}{%
\small
\setlength\tabcolsep{4pt} 
\begin{tabular}{l|c c|c c|c c|c c|c c}
\hline
\multirow{ 2}{*}{Model} & \multicolumn{2}{c|}{KP20k} & \multicolumn{2}{c|}{Inspec} & \multicolumn{2}{c|}{Krapivin} & \multicolumn{2}{c|}{NUS} & \multicolumn{2}{c}{SemEval} \\
\cline{2-11}
& F1@M & F1@5 & F1@M & F1@5 & F1@M & F1@5 & F1@M & F1@5 & F1@M & F1@5 \\ 
\hline
\multicolumn{11}{l}{Present Keyphrase Generation} \\
\hline
catSeq &  0.367 & 0.291 & 0.262 & 0.225 & 0.354 & 0.269 & 0.397 & 0.323 & 0.283 & 0.242 \\
catSeqD & 0.363 & 0.285 & 0.263 & 0.219 & 0.349 & 0.264 & 0.394 & 0.321 & 0.274 & 0.233 \\
catSeqCorr  & 0.365 & 0.289 & 0.269 & 0.227 & 0.349 & 0.265 & 0.390 & 0.319 & 0.290 & 0.246 \\
catSeqTG  & 0.366 & 0.292 & {\bf 0.270} & {\bf 0.229} & {\bf 0.366} & \underline{\bf 0.282} & 0.393 & 0.325 & 0.290 & 0.246 \\
Transformer &  0.368 & 0.291 & 0.264 & 0.225 & 0.356 & 0.274 & 0.405 & 0.328 & 0.288 & 0.245 \\
SEG-Net & \underline{\bf 0.379} & \underline{\bf 0.311} & 0.265 & 0.216 & {\bf 0.366} & 0.276 & \underline{\bf 0.461} & \underline{\bf 0.396} & \underline{\bf 0.332} & \underline{\bf 0.283} \\
\hline
\multicolumn{11}{l}{Absent Keyphrase Generation} \\
\hline
catSeq & 0.032 & 0.015 & 0.008 & 0.004 & 0.036 & 0.018 & 0.028 & 0.016 & 0.028 & 0.020 \\
catSeqD & 0.031 & 0.015 & 0.011 & 0.006 & 0.037 & 0.018 & 0.024 & 0.015 & 0.024 & 0.016 \\
catSeqCorr  & 0.032 & 0.015 & 0.009 & 0.005 & {\bf 0.038} & {\bf 0.020} & 0.024 & 0.014 & 0.026 & 0.018 \\
catSeqTG  & 0.032 & 0.015 & 0.011 & 0.005 & 0.034 & 0.018 & 0.018 & 0.011 & 0.027 & 0.019 \\
Transformer & 0.031 & 0.015 & 0.009 & 0.005 & {\bf 0.038} & {\bf 0.020} & 0.028 & 0.016 & 0.029 & 0.020 \\
SEG-Net & \underline{\bf 0.036} & \underline{\bf 0.018} & \underline{\bf 0.015} & {\bf 0.009} & 0.036 & 0.018 & \underline{\bf 0.036} & \underline{\bf 0.021} & {\bf 0.030} & {\bf 0.021} \\
\hline
\end{tabular}
}
\caption{
Results of keyphrase prediction on the scientific benchmarks.
The bold-faced and underline values indicate the best and statistically significantly better (by paired bootstrap test, $p < 0.05$) performances across the board.
}
\label{result:main}
\end{table*}

\subsection{Baseline Models and Evaluation Metrics}
We compare the performance of SEG-Net with four state-of-the-art neural generative methods, catSeq \cite{yuan-etal-2020-one}, catSeqD \cite{yuan-etal-2020-one}, catSeqCorr \cite{chen-etal-2018-keyphrase}, and catSeqTG \cite{chen2019guided}.
In addition, we consider the vanilla Transformer \cite{vaswani2017attention} as a baseline.
The catSeq, catSeqCorr and catSeqTG models are known as CopyRNN \cite{meng-etal-2017-deep}, CorrRNN \cite{chen-etal-2018-keyphrase} and TGNet \cite{chen2019guided} respectively.
CopyRNN, CorrRNN or TGNet generates one keyphrase in a sequence-to-sequence fashion and use beam search to generate multiple keyphrases.
In contrast, following \citet{chan-etal-2019-neural}, we concatenate all the keyphrases into one output sequence using a special delimiter $\langle sep \rangle$, and use greedy decoding during inference.
We train all the baselines using maximum-likelihood objective.
We use the publicly available implementation of these baselines\footnote{https://github.com/kenchan0226/keyphrase-generation-rl} in our experiment.

To measure the accuracy of the sentence-selector, we use averaged F1 score (macro).
We also compute precision and recall to compare the performance of the sentence-selector with a baseline.
While in SEG-Net, we select up to $N$ predicted salient sentences, in the baseline method, the first $N$ sentences are selected from the target document so that their total length does not exceed a predefined word limit (200 words).
In keyphrase generation, the accuracy is typically computed by comparing the top $k$ predicted keyphrases with the ground-truth keyphrases.
We follow \citet{chan-etal-2019-neural} to perform evaluation and report F1@M and F1@5 for all the baselines and our models.

\subsection{Implementation Details}

\paragraph{Hyper-parameters}
We use a fixed vocabulary of the most frequent $|V| = 50,000$ words in both sentence-selector and extractor-generator.
We set $d_{model}=512$ for all the embedding vectors.
We set $L=6, h=8, d_k=64, d_v=64, d_{ff}=2,048$ in Transformer across all our models.
We detail the hyper-parameters in Table \ref{table:hyperparameters} in Appendix.

\paragraph{Training}
We perform grid search for $\beta$ over [0.4, 0.5, 0.6] on the dev set and found $\beta=0.5$ results in the best performance.
Loss weights for positive samples $\omega$ are set to 0.7 and 2.0 during selector and extractor training.\footnote{The values are chosen by simply computing the fraction of the positive and negative samples.}
We train all our models using Adam \cite{kingma2014adam} with a batch size of 80 and a learning rate of $10^{-4}$.
During training, we use dropout and gradient clipping.
We halve the learning rate when the validation performance drops and stop training if it does not improve for five successive iterations.
We train the sentence-selector and extractor-generator modules for a maximum of 15 and 25 epochs, respectively. 
Training the modules takes roughly 10 and 25 hours on two GeForce GTX 1080 GPUs, respectively.

\paragraph{Decoding}
The absent keyphrases are generated as a concatenated sequence of words.
Hence, unlike prior works \cite{meng-etal-2017-deep,chen-etal-2018-keyphrase,chen2019guided,chen-etal-2019-integrated,zhao-zhang-2019-incorporating}, we use greedy search as the decoding algorithm during testing, and we force the decoder never to output the same trigram more than once to avoid repetitions in the generated keyphrases.
This is accomplished by not selecting the word that would create a trigram already exists in the previously decoded sequence.
It is a well-known technique utilized in text summarization \cite{paulus2017deep}.

We provide details about model implementations and references in Appendix for reproducibility.

\begin{table}[!t]
\centering
\resizebox{\linewidth}{!}{%
\small
\setlength\tabcolsep{3pt} 
\begin{tabular}{l|c c|c c}
\hline
\multirow{ 2}{*}{Model} & \multicolumn{2}{c|}{KPTimes} & \multicolumn{2}{c}{In-house} \\
\cline{2-5}
& F1@M & F1@5 & F1@M & F1@5 \\ 
\hline
\multicolumn{5}{l}{Present Keyphrase Generation} \\
\hline
catSeq & 0.453 & 0.295 & 0.255 & 0.102 \\
catSeqD & 0.456 & 0.299 & 0.252 & 0.100 \\
catSeqCorr & 0.457 & 0.302 & 0.247 & 0.100 \\
catSeqTG & 0.465 & 0.310 & 0.260 & 0.103 \\
Transformer & 0.451 & 0.296 & 0.258 & 0.111 \\
SEG-Net & \underline{\bf 0.481} & \underline{\bf 0.367} & \underline{\bf 0.298} & \underline{\bf 0.161} \\
\hline
\multicolumn{5}{l}{Absent Keyphrase Generation} \\
\hline
catSeq & 0.227 & 0.157 & 0.041 & 0.020 \\
catSeqD & 0.225 & 0.158 & 0.037 & 0.019 \\
catSeqCorr & 0.225 & 0.158 & 0.037 & 0.019 \\
catSeqTG & 0.227 & 0.155 & 0.037 & 0.018 \\
Transformer & 0.218 & 0.148 & 0.042 & 0.020 \\
SEG-Net & \underline{\bf 0.237} & \underline{\bf 0.169} & \underline{\bf 0.047} & \underline{\bf 0.024} \\
\hline
\end{tabular}
}
\caption{
Keyphrase prediction results on the two web domain benchmarks.
The bold-faced values and $^\dagger$ indicate the best and statistically significantly better (by paired bootstrap test, $p<0.05$) performances.
}
\label{result:main_web}
\vspace{2mm}
\end{table}

\begin{table}[!ht]
\centering
\resizebox{\linewidth}{!}{%
\small
\begin{tabular}{l|c@{\hskip 0.1in} c| c@{\hskip 0.1in} c}
\hline
\multirow{ 2}{*}{Model} & \multicolumn{2}{c|}{Present} & \multicolumn{2}{c}{Absent} \\ 
\cline{2-5}
& MAE & Avg. \# & MAE & Avg. \# \\ 
\hline
Oracle &  0.000 & 2.837 & 0.000 & 2.432 \\
\hline
catSeq & 2.271 & 3.781 & 1.943 & 0.659 \\
catSeqD & 2.225 & 3.694 & 1.961 & 0.629 \\
catSeqCorr & 2.292 & 3.790 & 1.914 & 0.703 \\
catSeqTG & 2.276 & 3.780 & 1.956 & 0.638 \\
\hline
SEG-Net & \textbf{2.185} & 3.796 & \textbf{1.324} & 1.140 \\
\hline
\end{tabular}
}
\caption{
Evaluation on predicting the correct number of keyphrases on the KP20k dataset. MAE stands for mean absolute error and ``Avg. \#'' indicates the average number of generated keyphrases per document. Oracle is a model that generates the ground-truth keyphrases.
}
\label{result:correct_kp}
\end{table}

\section{Results}

We compare our proposed model SEG-Net with the baselines on the scientific and web domain datasets. We present the experiment results in Table \ref{result:main} and \ref{result:main_web}.

\paragraph{Present keyphrase prediction}
From the results, it is evident that SEG-Net outperforms all the baseline methods by a significant margin ($p < 0.05$, $t$-test) in 3 out of 5 scientific datasets and both web domain datasets.
Unlike the baseline methods, SEG-Net extracts the present keyphrases from the salient sentences, contributing most to the performance improvement.
In the Krapivin dataset, the performance is on par, while in the Inspec dataset, SEG-Net performs worst in terms of F1@5.
The perofrmance drop is explainable as Inspect dataset consists of shorter documents (see the average lengths in Table \ref{table:statitics}). 
In NUS and SemEval datasets, the performance improvements are noteworthy; 5.6 and 4.2 F1@M points over the second-best method.
The number of ground truth keyphrases in those two datasets are higher than other scientific datasets, and extracting present keyphrases boosts the performance (more discussion in \cref{sec:ablation}).
SEG-Net significantly improves the web domain datasets (3.1 F1@5 points in KPTimes and 5.0 F1@5 points in In-house datasets) over the best baseline methods, catSeqTG, and Transformer, respectively.

\paragraph{Absent keyphrase prediction}
Unlike present phrases, absent phrases do not appear exactly in the target document. Hence, predicting them is more challenging and requires a comprehensive understanding of the underlying document semantic.
From Table \ref{result:main} and \ref{result:main_web}, we see that SEG-Net correctly generates more absent keyphrases than the baselines on all the experimental datasets, except Krapivin.
To our surprise, SEG-Net results in a large performance improvement (1.0 points in terms of F1@M) in the KPTimes dataset.
We suspect that in KPTimes dataset, the target absent keyphrases are semantically associated with different segments of the document and thus generating such keyphrases from the salient sentences results in larger improvements.
Overall, the absent phrase prediction results indicate that SEG-Net is capable of understanding the underlying document semantic better than the baseline methods.

\paragraph{Number of generated keyphrases}
Generating an accurate number of keyphrases indicates models' understanding of the documents' semantic.
A small number of phrase predictions demonstrate a model's inability to identify all the key points; over generation implies a model's wrong understanding of the crucial points.
Hence, we compare SEG-Net with all the baseline approaches for predicting the appropriate number of phrases.
We measure the mean absolute error (MAE) between the number of generated keyphrases and the number of ground-truth keyphrases \cite{chan-etal-2019-neural}.
The results for KP20k are presented in Table \ref{result:correct_kp}.
The lower MAEs for SEG-Net indicate it better understands documents' semantic.
However, in the KPTimes dataset, we observe SEG-Net predicts more present keyphrases than the baselines (see Table 3 in Appendix).
This is due to the extractive nature of SEG-Net, and documents having more closely related keyphrases (e.g., SEG-Net predicts ground-truth keyphrases: ``Google'', ``Apple'' with other relevant keyphrases: ``line'', ``Amazon.com''. See qualitative examples provided in Appendix).
Therefore, we suggest future works to consider the dataset nature while judging models in this respect.

\begin{table}[!t]
\centering
\resizebox{\linewidth}{!}{%
\small
\setlength\tabcolsep{3pt} 
\begin{tabular}{l|l|r r|r r}
\hline
& & \multicolumn{2}{c|}{Present} & \multicolumn{2}{c}{Absent} \\
\cline{3-6}
& & F1@M & F1@5 & F1@M & F1@5 \\ 
\hline
\multirow{4}{*}{\rotatebox[origin=c]{90}{KP20k}} & SEG-Net & 0.379 & 0.311 & 0.036 & 0.018 \\ \cline{2-6}
& w/o DEG & $-$0.004 & $-$0.005 & $-$0.001 & 0.000 \\
& w/o SSS & $-$0.008 & $-$0.014 & $+$0.001 & $+$0.001 \\
& w/o LCA & $+$0.001 & 0.000 & $-$0.004 & $-$0.002 \\
\hline
\multirow{4}{*}{\rotatebox[origin=c]{90}{NUS}} & SEG-Net & 0.461 & 0.396 & 0.036 & 0.021 \\ \cline{2-6}
& w/o DEG & $-$0.028 & $-$0.031 & $-$0.002 & $-$0.001 \\
& w/o SSS & $-$0.044 & $-$0.052 & 0.000 & $+$0.001 \\
& w/o LCA & $-$0.004 & $-$0.002 & $-$0.004 & $-$0.003 \\
\hline
\multirow{4}{*}{\rotatebox[origin=c]{90}{SemEval}} & SEG-Net & 0.332 & 0.283 & 0.030 & 0.021 \\ \cline{2-6}
& w/o DEG & $-$0.010 & $-$0.009 & 0.000 & 0.000 \\
& w/o SSS & $-$0.035 & $-$0.032 & $-$0.001 & $-$0.001 \\
& w/o LCA & $-$0.002 & $-$0.002 & $-$0.001 & $-$0.001 \\
\hline
\multirow{4}{*}{\rotatebox[origin=c]{90}{KPTimes}} & SEG-Net & 0.428 & 0.367 & 0.187 & 0.119 \\ \cline{2-6}
& w/o DEG & $-$0.022 & $-$0.059 & $-$0.008 & $-$0.005 \\
& w/o SSS & $-$0.038 & $-$0.067 & $-$0.028 & $-$0.022 \\
& w/o LCA & $-$0.005 & $-$0.010 & $-$0.030 & $-$0.018 \\
\hline
\end{tabular}
}
\caption{
Ablation on SEG-Net without decoupling extraction and generation (DEG), salient sentence selection (SSS), and layer-wise coverage attention (LCA).
We preclude one design choice at a time.
}
\label{ablation:result_main}
\end{table}

\section{Analysis}
\label{sec:ablation}

The differences between the Transformer baseline and SEG-Net are (1) decoupling keyphrase extraction and generation, (2) use salient sentences for keyphrase prediction, and (3) layer-wise coverage attention.
We perform ablation on the three design choices and present the results in Table \ref{ablation:result_main}.

\paragraph{Decoupling extraction and generation}
SEG-Net extracts present keyphrases and generates absent keyphrases as suggested in \citet{chen-etal-2019-integrated} with a difference in the extractor.
SEG-Net employs a 3-way classifier (to predict BIO tags) that enables consecutive present keyphrases extraction.
The ablation study shows that separating extraction and generation boosts present keyphrase prediction (as much as 2.8, 1.0, and 2.2 F1@M points in NUS, SemEval, and KPTimes datasets, respectively).

\begin{figure}
\centering
\subfloat[Present keyphrase]{
    \centering
    \includegraphics[width=0.95\linewidth]{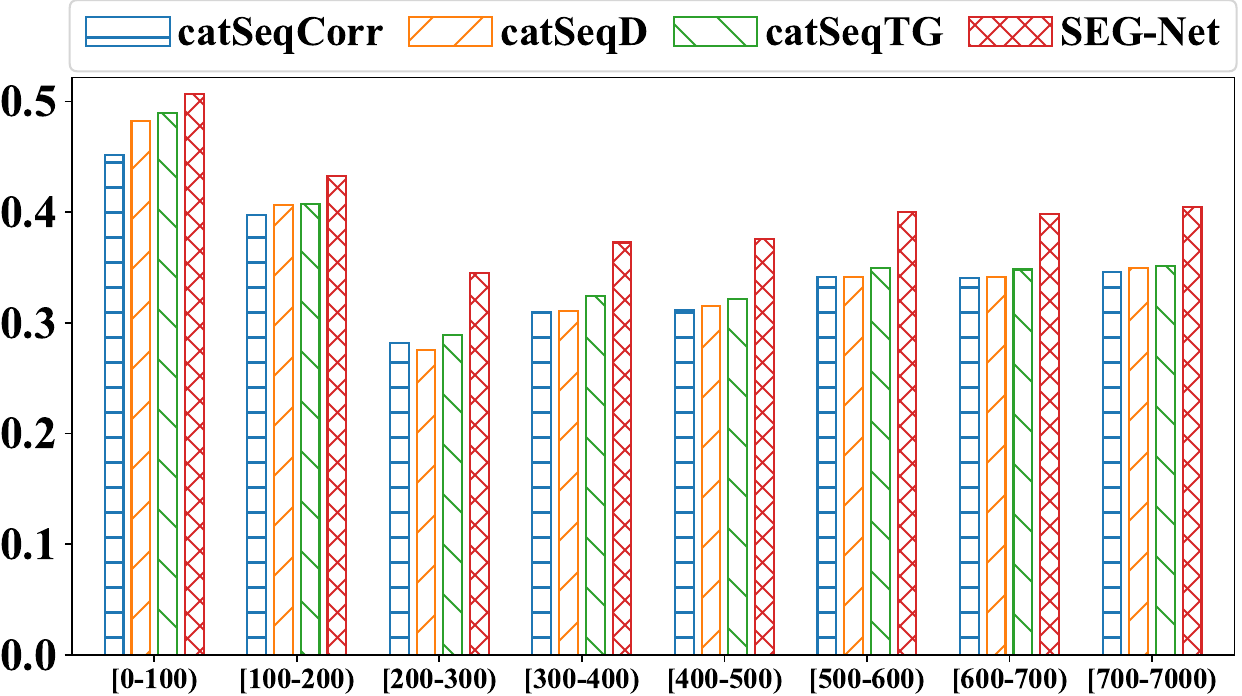}
    \label{subfig:present_kp} 
}
\\
\subfloat[Absent keyphrase]{
    \centering
    \includegraphics[width=0.95\linewidth]{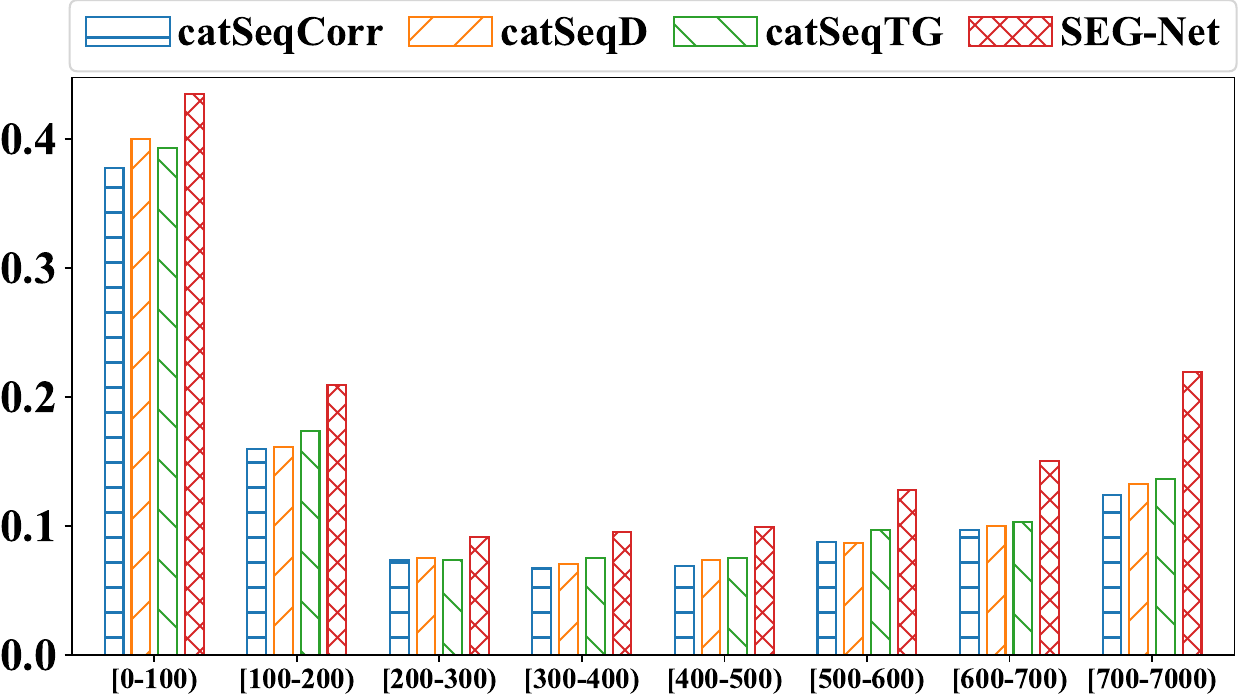}
    \label{subfig:absent_kp} 
}
\caption{
Test performance of different models on KPTimes dataset.
The x-axis and y-axis indicates document length (\# words) and F1@M score, respectively.
} 
\label{fig:kptimes_anal} 
\end{figure}

\paragraph{Salient sentence selection}
One of SEG-Net's key contributions is the sentence-selector that identifies the salient sentences to minimize the risk of missing critical points due to truncating long target documents (e.g., web documents).
The contribution of sentence-selector in present keyphrase prediction is evident from the ablation study.
The impact of using salient sentences to generate absent keyphrases is significant for the web domain datasets (e.g., 2.8 F1@M points in KPTimes). 
We show the performances on KPTimes test documents with different length in Figure 
\ref{fig:kptimes_anal} and the results suggest that SEG-Net improves absent keyphrase prediction significantly for longer documents, and we credit this to the sentence selector.
The selector's accuracy on the KP20k and KPTimes datasets are 78.2 and 73.7 in terms of (macro) F1 score.
We evaluate SEG-Net by providing the ground-truth salient sentences to quantify the improvement achievable with a perfect sentence-selector.
We found that the present keyphrase prediction performance would have increased by 3.2 and 4.1 F1@M points with a perfect sentence-selector.

We compare the sentence selector with the baselines that select the first $N$ sentences from the target document, and the results are presented in Table \ref{result:selector}.
SEG-Net's selector has a higher precision that indicates it processes input texts with more salient sentences.
On the other hand, the recall is substantially lower for the scientific domain due to false-negative predictions.
Our experiments suggest that salient sentence selection positively impacts and has additional room for improvement.


\begin{table}[!t]
\centering
\resizebox{\linewidth}{!}{%
\small
\begin{tabular}{l|c|c|c|c}
\hline
\multirow{2}{*}{Dataset}  & \multicolumn{2}{c|}{SEG-Net} & \multicolumn{2}{c}{Baseline} \\
\cline{2-5}
& Prec. & Recall & Prec. & Recall \\
\hline
\multicolumn{5}{l}{Scientific Domain} \\
\hline
KP20k & 84.5 & 86.3 & 75.1 & 95.0 \\
Inspec & 95.0 & 82.6 & 87.1 & 98.8 \\
Krapivin & 85.5 & 85.3 & 75.8 & 95.1 \\
NUS & 91.8 & 81.0 & 78.1 & 92.0  \\
SemEval & 97.1 & 75.7 & 83.5 & 90.3 \\
\hline
\multicolumn{5}{l}{Web Domain} \\
\hline
KPTimes & 81.7 & 44.9 & 73.0 & 45.7 \\
In-house & 82.9 & 49.1 & 66.8 & 51.3 \\
\hline
\end{tabular}
}
\caption{
Precision and recall computed by selecting $N$ \emph{predicted} salient sentences in SEG-Net, and the \emph{first} $N$ sentences from the target documents in the baselines.
We set $N$ for each target document so that the total length of the selected sentences does not exceed a limit of 200 words.
It is important to note that the baseline recall is close to 100.0 for the scientific domain datasets because the average length of the target documents from that domain is closer to 200 words.
}
\label{result:selector}
\end{table}

\paragraph{Layer-wise coverage attention}
The ablation study shows the positive impact of the layer-wise coverage attention in SEG-Net. The improvement in absent keyphrase generation for the KPTimes dataset (3.0 F1@M points) is significant, while it is relatively small in other experiment datasets. 
We hypothesize that the coverage attention helps when keyphrases summarize concepts expressed in different segments of a long document.
We confirm our hypothesis by observing the performance trend with and without the coverage attention mechanism (we observe a similar trend as in Figure \ref{fig:kptimes_anal}).

We provide additional experiment results and qualitative examples in Appendix.

\section{Related Work}

Keyphrase extraction approaches identify important phrases that appear in a document.
The existing approaches generally work in two steps.
First, they select a set of candidate keyphrases based on heuristic rules \cite{hulth-2003-improved,medelyan2008topic,liu-etal-2011-automatic,wang2016ptr}.
The selected keyphrases are scored as per their importance in the second step,
which is computed by unsupervised ranking approaches \cite{wan2008single,grineva2009extracting} or supervised learning algorithms \cite{hulth-2003-improved,witten2005kea,medelyan-etal-2009-human,nguyen2007keyphrase,lopez-romary-2010-humb}. 
Finally, the top-ranked candidates are returned as the keyphrases. Another pool of extractive solutions follows a sequence tagging approach  \cite{luan-etal-2017-scientific,zhang-etal-2016-keyphrase,gollapalli2017incorporating,gollapalli2014extracting}. 
However, the extractive solutions are only able to predict the keyphrases that appear in the document and thus fail to predict the absent keyphrases.

Keyphrase generation methods aim at predicting both the present and absent phrases.
\citet{meng-etal-2017-deep} proposed the first generative model, known as CopyRNN, which is composed of attention \cite{bahdanau2014neural,luong-etal-2015-effective} and copy mechanism \cite{gu-etal-2016-incorporating,see-etal-2017-get}. 
Multiple extensions of CopyRNN were proposed in subsequent works \cite{chen-etal-2018-keyphrase, chen2019guided}.
Different from these approaches, \citet{zhang2017deep} proposed CopyCNN that utilizes convolutional neural network (CNN) \cite{kim2014convolutional} to form sequence-to-sequence architecture.
However, these generation methods are trained to predict \emph{one} keyphrase from the target document.
In contrast, \citet{yuan-etal-2020-one} proposed to concatenate all the ground-truth keyphrases and train models to generate them as one output sequence.

Other noteworthy approaches in literature utilize data from external source \cite{chen-etal-2019-integrated},
syntactic supervision \cite{zhao-zhang-2019-incorporating}, semi-supervised learning \cite{ye-wang-2018-semi}, reinforcement learning \cite{chan-etal-2019-neural}, adversarial training \cite{swaminathan-etal-2020-preliminary}, unlikelihood training \cite{bahuleyan-el-asri-2020-diverse} to improve keyphrase generation.

\section{Conclusion}
This paper presents SEG-Net, a keyphrase generation model that identifies the salient sentences in a target document to utilize maximal information for keyphrase prediction.
In SEG-Net, we incorporate a novel layer-wise coverage attention to cover all the critical points in a document and diversify the present and absent keyphrases.
We evaluate SEG-Net on seven benchmarks from scientific and web documents, and the experiment results demonstrate SEG-Net's effectiveness over the state-of-the-art methods on both domains.


\bibliography{anthology,acl2021}
\bibliographystyle{acl_natbib}

\clearpage
\appendix











\twocolumn[{%
 \centering
 \Large\bf Supplementary Material: Appendices \\ [20pt]
}]

\begin{table}[t]
\centering
\resizebox{\linewidth}{!}{%
\begin{tabular}{l|c c| c c}
\hline
\multirow{2}{*}{Model} & \multicolumn{2}{c|}{Present} & \multicolumn{2}{c}{Absent} \\ 
\cline{2-5}
& F1@M & F1@5 & F1@M & F1@5 \\
\hline
catSeq & 0.376 & 0.298 & 0.034 & 0.016 \\
catSeqD  & 0.372 & 0.293 & 0.033 & 0.016 \\
catSeqCorr & 0.375 & 0.300 & 0.034 & 0.016 \\
catSeqTG & 0.374 & 0.302 & 0.033 & 0.016 \\
\hline
SEG-Net & \textbf{0.390} & \textbf{0.326} & \textbf{0.042} & \textbf{0.021} \\
\hline
\end{tabular}
}
\caption{
Test set results on the KP20k dataset with ``name variations'' as proposed in \citet{chan-etal-2019-neural}.
}
\label{result:name_variation}
\end{table}

\begin{table}[!t]
\centering
\begin{tabular}{l|l|c c}
\hline
& \multirow{ 2}{*}{Input features} & \multicolumn{2}{c}{Present} \\ 
\cline{3-4}
& & F1@M & F1@5 \\ 
\hline
\multirow{3}{*}{\rotatebox[origin=c]{90}{KP20k}} & SEG-Net & 0.379 & 0.311 \\
& w/o Character Emb. & 0.376 & 0.309 \\
& w/o Segment Emb. & 0.378 & 0.310  \\ \hline
\multirow{3}{*}{\rotatebox[origin=c]{90}{KPTimes}} & SEG-Net & 0.481 & 0.367 \\
& w/o Character Emb. & 0.462 & 0.332 \\
& w/o Segment Emb. & 0.475 & 0.365  \\ \hline
\multirow{3}{*}{\rotatebox[origin=c]{90}{In-house}} & SEG-Net & 0.298 & 0.161 \\
& w/o Character Emb. & 0.284 & 0.152 \\
& w/o Segment Emb. & 0.295 & 0.159  \\
\hline
\end{tabular}
\caption{Impact of different embeddings at the input layer in SEG-Net.
}
\label{ablation:word_rep}
\end{table}
\begin{table}[t]
\centering
\resizebox{\linewidth}{!}{%
\small
\begin{tabular}{l|c@{\hskip 0.1in} c|c@{\hskip 0.1in} c}
\hline
\multirow{2}{*}{Model} & \multicolumn{2}{c|}{Present} & \multicolumn{2}{c}{Absent} \\ 
\cline{2-5}
& F1@M & F1@5 & F1@M & F1@5 \\
\hline
\multicolumn{5}{l}{KP20k} \\
\hline
catSeqTG & {\bf 0.386} & {\bf 0.321} & 0.050 & 0.027 \\ 
SEG-Net & 0.380 & 0.311 & {\bf 0.052} & {\bf 0.030} \\
\hline
\multicolumn{5}{l}{KPTimes} \\
\hline
catSeqTG & {\bf 0.481} & 0.318 & 0.238 & 0.174 \\ 
SEG-Net & 0.475 & {\bf 0.358} & {\bf 0.245} & {\bf 0.181} \\
\hline
\end{tabular}
}
\caption{
Test set results after fine-tuning the models via RL as proposed in \citet{chan-etal-2019-neural}.
}
\label{result:rl_results}
\end{table}

\begin{table}[!ht]
\centering
\resizebox{\linewidth}{!}{%
\small
\begin{tabular}{l|c@{\hskip 0.1in} c| c@{\hskip 0.1in} c}
\hline
\multirow{ 2}{*}{Model} & \multicolumn{2}{c|}{Present} & \multicolumn{2}{c}{Absent} \\ 
\cline{2-5}
& MAE & Avg. \# & MAE & Avg. \# \\ 
\hline
Oracle &  0.000 & 3.054 & 0.000 & 1.978 \\
\hline
catSeq & 1.437 & 2.141 & 1.297 & 2.397 \\
catSeqD & 1.431 & 2.193 & 1.369 & 2.523 \\
catSeqCorr & 1.469 & 2.277 & 1.373 & 2.520 \\
catSeqTG & 1.378 & 2.309 & 1.284 & 2.342 \\
\hline
SEG-Net & 2.209 & 4.650 & 1.291 & 2.196 \\
\hline
\end{tabular}
}
\caption{
Evaluation on predicting the correct number of keyphrases on the KPTimes dataset. MAE stands for mean absolute error and ``Avg. \#'' indicates the average number of generated keyphrases per document. Oracle is a model that generates the ground-truth keyphrases.
}
\label{result:mae_kptimes}
\end{table}

\section{Additional Ablation Study}

\paragraph{Variation of named entities}
A keyphrase can be expressed in different ways, such as ``solid state drive'' as ``ssd'' or ``electronic commerce'' as ``e commerce'' etc.
A model should receive credit if it generates any of those variations.
Hence, \citet{chan-etal-2019-neural} aggregated name variations of the ground-truth keyphrases from the KP20k evaluation dataset using the Wikipedia knowledge base.
We evaluate our model on that enriched evaluation set, and the experimental results are listed in Table \ref{result:name_variation}.
We observed that although SEG-Net extracts the present keyphrases, it can predict present phrases with variations such as ``support vector machine'' and ``svm'' if they co-exist in the target document.

\paragraph{Impact of embedding features}
The embedding layer of extractor-generator learns four different embedding vectors: word embedding, position embedding, character-level embedding, and segment embedding that are element-wise added.
We remove character embedding and segment embedding and observe slight performance drop in present keyphrase prediction.
The results are presented in Table \ref{ablation:word_rep}.
The character embeddings are employed as we limit the vocabulary to the most frequent $V$ words.
During our preliminary experiment, we observed that character embeddings have a notable impact in the web domain, where the actual vocabulary size can be large.
The addition of segment embedding is also helpful, specially the sentence-selector may predict salient sentences from any part of the document. We hypothesize that the sentence index guides the self-attention mechanism in the extractor-generator.

\paragraph{Fine-tuning via Reinforcement Learning}
Following \citet{chan-etal-2019-neural}, we apply reinforcement learning (RL) to fine-tune the extractor-generator module of SEG-Net on absent keyphrase generation. 
As we can see from Table \ref{result:rl_results}, due to RL fine-tuning, the absent keyphrase generation improves significantly, which corroborates with the findings of \citet{chan-etal-2019-neural}.
While fine-tuning catSeqTG model via RL helps present keyphrase generation in KP20k, it does not help in KPTimes dataset.
Since SEG-Net extracts the present keyphrases, their predictions do not benefit from the RL fine-tuning step (instead, performance drops slightly).

\begin{table}[t]
\centering
\resizebox{\linewidth}{!}{%
\begin{tabular}{l|c}
\hline
Vocabulary size, $|V|$ & 50,000 \\
\# CNN filters & 512 1D \\
Model size, $d_{model}$ & 512 \\
encoder layers & 6 \\
decoder layers & 6 \\
$h$, $d_k$, $d_v$, $d_{ff}$ & 8, 64, 64, 2048 \\
dropout & 0.2 \\
optimizer & Adam \\
learning rate & 0.0001 \\
learning rate decay & 0.5 \\
batch size & 80 \\
Maximum gradient norm & 1.0 \\
\# Params (sentence-selector) & 41.6M \\
\# Params (extractor-generator) & 54.2M \\
\hline
\end{tabular}
}
\caption{
Hyper-parameters used to train SEG-Net. We use the same setup for the Transformer model.
}
\label{table:hyperparameters}
\end{table}

\section{Evaluation Metrics}

We want to draw attention to a crucial detail about the evaluation metric setup. 
Due to differences in post-processing before computing the evaluation metric values, the reported scores in papers differ.
Recent works in literature mostly follow either evaluation metric implementation from \citet{chan-etal-2019-neural} or \citet{yuan-etal-2020-one}.
Both works have shared their implementation publicly available, and we use the implementation of \citet{chan-etal-2019-neural}.

We reported F1@5 and F1@M scores in this work, where M denotes the number of predicted keyphrases. We also compute F1@10 and F1@O, where O represents the number of ground truth keyphrases, and the results are presented in Table \ref{result:extra}.
Many prior works have reported R@10 and R@50 for absent phrase generation. To compute R@50, we need to perform beam decoding to generate many keyphrases, typically more than 200 \cite{yuan-etal-2020-one}. In our opinion, generating hundreds of keyphrases from a document does not truly reflect the models' ability in understanding document semantic. Therefore, we do not prefer to assess models' ability in terms of R@50 metrics.

\begin{table}[!t]
\centering
\resizebox{\linewidth}{!}{%
\small
\begin{tabular}{l|c@{\hskip 0.1in} c|c@{\hskip 0.1in} c}
\hline
\multirow{ 2}{*}{Model} & \multicolumn{2}{c|}{Present} & \multicolumn{2}{c}{Absent} \\
\cline{2-5}
& F1@10 & F1@O & F1@10 & F1@O \\ \hline
KP20k & 0.201 & 0.350 & 0.012 & 0.027 \\
Inspec & 0.140 & 0.201 & 0.005 & 0.011 \\
Krapivin & 0.172 & 0.315 & 0.011 & 0.025 \\
NUS & 0.270 & 0.378 & 0.013 & 0.022 \\
SemEval & 0.199 & 0.258 & 0.014 & 0.018 \\
KPTimes & 0.244 & 0.464 & 0.122 & 0.208 \\
In-house & 0.094 & 0.282 & 0.014 & 0.035 \\
\hline
\end{tabular}
}
\caption{
Present and absent keyphrase prediction results on the experiment datasets.
}
\label{result:extra}
\end{table}

\section{Qualitative Analysis}
We provide a few qualitative examples in Figure \ref{table:qual_example}.

\begin{figure*}[ht!]
\centering
\resizebox{\linewidth}{!}{%
\def\arraystretch{1.5}%
\begin{tabular}{ p{0.99\linewidth}}
\hline
\hline
\textbf{Title:}
\href{https://www.japantimes.co.jp/news/2017/07/10/reference/smart-speakers-powered-voice-agents-seen-ushering-era-ai/}{\color{blue}smart speakers powered by voice agents seen ushering in era of ai} \\
\hline
\textbf{Article:}
major tech firms have been keen to sell speakers equipped with voice - based artificial intelligence agents recently . [EOS] the debuts of smart speakers are seen as the prelude to an ai era , ushering in a new technological age in which virtual assistants are expected to become as ubiquitous as smartphones , allowing people to connect to the internet by voice with greater ease . [EOS] whether these speakers will really take off and whether the technology will be popular in japan remain to be seen . [EOS] the following questions and answers explore these issues as well as why ai speakers are creating a buzz and what will be the role of japanese firms in this field . [EOS] what makes ai speakers special ? they look like normal portable home speakers , but one big difference is that they communicate with users verbally . [EOS] users can tell the speakers to play music , search the internet , pull up weather forecasts , send text messages , make phone calls and perform other daily tasks . [EOS] 
... (truncated) \\
\hline
\textbf{[catSeq]} 
smartphones ; artificial intelligence ; science and technology   \\
\hline
\textbf{[SEG-Net]} 
smart speakers ; smartphones ; {\color{red}ai} ; japan ;  speakers ; {\color{red}google} ; {\color{red}apple} ; computers and the internet ; tech industry \\
\hline
\textbf{[Ground-truth]} 
{\color{red}google} ; {\color{red}apple} ; line ; {\color{red}ai} ; amazon.com ; iot \\
\hline
\hline
\textbf{Title:}
\href{https://www.japantimes.co.jp/news/2014/08/29/national/science-health/basic-information-on-dengue-fever-2/}{\color{blue}how much do you know about dengue fever ?}
  \\
\hline
\textbf{Article:}
the health ministry has confirmed the first domestic dengue fever case in japan in nearly 70 years . [EOS] a saitama prefecture teen girl was found wednesday to have contracted the virus through a mosquito in japan , followed by news that two more people — a man and a woman in tokyo — have also been infected . [EOS] more than 200 dengue cases are reported in japan each year , but those are of patients who contracted dengue virus abroad . [EOS] the world health organization estimates the number of infections across the globe to be 50 million to 100 million per year . [EOS] while the news has led to widespread fears that a pandemic outbreak might have arrived , experts are quick to deny such a scenario , while offering some advice on what measures people can take to minimize their exposure . [EOS] following are some basic questions and answers regarding the infectious disease and measures that can be taken to prevent infection . [EOS] what is dengue fever and what causes it ? dengue fever is a tropical viral disease , also known as dengue hemorrhagic fever or break - bone fever , 
... (truncated) \\
\hline
\textbf{[catSeq]} 
{\color{red}dengue fever} ; japan; medicine and health  \\
\hline
\textbf{[SEG-Net]} 
{\color{red}dengue fever} ; japan ; dengue ; {\color{red}dengue virus} ; health organization ; mosquitoes ; vaccines immunization \\
\hline
\textbf{[Ground-truth]} 
{\color{red}dengue fever} ; world health organization ; {\color{red}dengue virus} ; infectious diseases \\
\hline
\hline
\textbf{Title:}
\href{https://www.japantimes.co.jp/news/2013/03/13/national/media-national/photo-report-foodex-japan-2013/}{\color{blue}photo report : foodex japan 2013} \\
\hline
\textbf{Article:}
foodex is the largest trade exhibition for food and drinks in asia , with about 70,000 visitors checking out the products presented by hundreds of participating companies . [EOS] i was lucky to enter as press ; otherwise , visitors must be affiliated with the food industry — and pay ¥ 5,000 — to enter . [EOS] the foodex menu is global , including everything from cherry beer from germany and premium mexican tequila to top - class french and chinese dumplings . [EOS] the event was a rare chance to try out both well - known and exotic foods and even see professionals making them . [EOS] in addition to booths offering traditional japanese favorites such as udon and maguro sashimi , there were plenty of innovative twists , such as dorayaki , a sweet snack made of two pancakes and a red - bean filling , that came in coffee and tomato flavors . [EOS]
... (truncated) \\
\hline
\textbf{[catSeq]} 
japan ; agriculture   \\
\hline
\textbf{[SEG-Net]} 
foodex japan ; {\color{red}foodex} ; food ; japan ; international trade and world market ; snack food \\
\hline
\textbf{[Ground-truth]} 
{\color{red}foodex} ; japanese food ; japan pulse \\
\hline
\end{tabular}
}
\vspace{-2mm}
\caption*{
}
\label{table:qual_example}
\end{figure*}

\begin{figure*}[ht!]
\centering
\resizebox{\linewidth}{!}{%
\def\arraystretch{1.5}%
\begin{tabular}{ p{0.99\linewidth}}
\hline
\hline
\textbf{Title:}
\href{https://www.japantimes.co.jp/news/2018/12/11/asia-pacific/social-issues-asia-pacific/majority-australian-women-sexually-harassed-work-survey/}{\color{blue}majority of australian women sexually harassed at work : survey} \\
\hline
\textbf{Article:}
kuala lumpur - two in three australian women have been sexually harassed at work , with the majority of cases unreported , according to a survey released on tuesday that highlighted challenges activists said prevent women from advancing in their careers . [EOS] some 64 percent of women and 35 percent of men said they had been harassed at their current or former workplace , according to the survey of over 9,600 people by the australian council of trade unions , the country ’s main group representing workers . [EOS] the majority of those surveyed said they were subjected to offensive behavior or unwanted sexual attention . [EOS] however only about a quarter of them made formal complaints , due to fears of repercussion , the survey found . [EOS] “ everyone should go to work free from the fear of harassment and unwanted sexual attention , ” the council ’s president , michele o’neil , said in a statement . [EOS] “ for many people — mainly women — today in australia this is not the reality . [EOS] our workplace laws have failed women who are experiencing harassment at work . [EOS] ” campaigners said sexual harassment creates a workplace environment that is discriminatory towards women , which can prevent them from moving forward in their careers . [EOS] `` sexual harassment in the workplace closes off women ’s opportunities and supports the attitudes that make violence more likely , ” merrindahl andrew , from the australian women against violence alliance , said by email . [EOS] australia was ranked 35 out of 144 countries in the world economic forum ’s 2017 gender gap index , up from 46 in 2016 due to greater female representation among legislators and managers . [EOS] although the global \# metoo movement has helped raised awareness about sexual harassment , the advocacy group plan international said the lack of strong policies and enforcement has discouraged victims from coming forward in australia . [EOS] 
... (truncated) \\
\hline
\textbf{[catSeq]} 
sexual harassment ; australian council ; {\color{red}australia} ; plan international ; [digit] presidential election ; michele e o’neil \\
\hline
\textbf{[SEG-Net]} 
workplace ; {\color{red}harassment} ; {\color{red}australia} ; sexual harassment ; women and girls ; women 's rights \\
\hline
\textbf{[Ground-truth]} 
{\color{red}australia} ; {\color{red}harassment} ; me too movement \\
\hline
\hline
\textbf{Title:}
\href{https://www.japantimes.co.jp/news/2019/03/15/national/science-health/google-team-led-japanese-engineer-breaks-record-calculating-pi-31-4-trillionth-digit/}{\color{blue}google team led by japanese engineer breaks record by calculating pi to the 31.4 trillionth digit} 
 \\
\hline
\textbf{Article:}
los angeles - google llc said thursday that a team led by engineer emma haruka iwao from japan has broken a guinness world record by calculating pi to the 31.4 trillionth digit , around 9 trillion more than the previous record set in 2016 . [EOS] the accomplishment , announced on the day dubbed `` pi day '' as its first three digits are 3.14 , was achieved by using google cloud infrastructure , the tech giant said . [EOS] iwao became fascinated with pi , an infinitely long number defined as the ratio of a circle ’s circumference to its diameter , when she was 12 years old . [EOS] `` when i was a kid , i downloaded a program to calculate pi on my computer , '' she said in a google blog post . [EOS] in college , one of her professors was daisuke takahashi of the university of tsukuba in ibaraki prefecture , then the record holder for calculating the most accurate value of pi via a supercomputer . [EOS] `` when i told him i was going to start this project , he shared his advice and some technical strategies with me , '' she said . [EOS] the groundbreaking calculation required 25 virtual google cloud machines , 170 terabytes of data and about 121 days to complete . [EOS] `` i 'm really happy to be one of the few women in computer science holding the record , and i hope i can show more people who want to work in the industry what 's possible , '' iwao said . [EOS] according to google , iwao calculated 31,415,926,535,897 digits , making it the first time the cloud has been used for a pi calculation of this magnitude . [EOS] \\
\hline
\textbf{[catSeq]} 
{\color{red}google} ; tv ; [digit] presidential election   \\
\hline
\textbf{[SEG-Net]} 
{\color{red}google} ; {\color{red}emma haruka iwao} ; japan ; google cloud ; computers and the internet ; tech industry \\
\hline
\textbf{[Ground-truth]} 
{\color{red}google} ; pi ; {\color{red}emma haruka iwao} ; mathematics \\
\hline
\end{tabular}
}
\caption{
Sample keyphrase predictions of catSeq and SEG-Net on KPTimes dataset (evaluation set). The highlighted keyphrases indicate a match with the ground truth keyphrases.
}
\label{table:qual_example}
\end{figure*}



\balance

\section{Reproducibility References}

\begin{itemize}
\setlength\itemsep{1pt}
\item We train and test the first four baseline models using their  \href{https://github.com/kenchan0226/keyphrase-generation-rl}{\color{blue}public implementation}. We use the Transformer implementation from \href{https://github.com/OpenNMT/OpenNMT-py}{\color{blue}OpenNMT} for catSeq (Transformer) and SEG-Net. 
\item We adopt the \href{https://github.com/neubig/util-scripts/blob/master/paired-bootstrap.py}{\color{blue}implementation} of paired bootstrap test script to perform significance test.
\item The preprocessed scientific article datasets are available \href{https://github.com/kenchan0226/keyphrase-generation-rl\#dataset}{\color{blue}here}. 
\item KPTimes dataset is available \href{https://github.com/ygorg/KPTimes}{\color{blue}here}. 
\end{itemize}


\end{document}